\documentclass{IOS-Book-Article}

\usepackage{mathptmx}
\usepackage{soul}\setuldepth{article}
\usepackage{subcaption}
\usepackage{todonotes}

\usepackage{hyperref}
\hypersetup{
    colorlinks=true,
    linkcolor=red,
    filecolor=magenta,      
    urlcolor=blue ,
    citecolor=green
    }
    
\newif\ifcomment
\commentfalse

\ifcomment
    \newcounter{MCRNumberOfComments}
    \stepcounter{MCRNumberOfComments}
    \newcommand{\mc}[1]{\textcolor{purple}{\small  [MC\#\arabic{MCRNumberOfComments}\stepcounter{MCRNumberOfComments}: #1]}}
    \newcounter{JVNumberOfComments}
    \stepcounter{JVNumberOfComments}
     \newcommand{\jv}[1]{\textcolor{blue}{\small  [JV\#\arabic{JVNumberOfComments}\stepcounter{JVNumberOfComments}: #1]}}
     
    \newcounter{AJNumberOfComments}
    \stepcounter{AJNumberOfComments}
     \newcommand{\aj}[1]{\textcolor{orange}{\small  [AJ\#\arabic{AJNumberOfComments}\stepcounter{AJNumberOfComments}: #1]}}
     
    \newcounter{SSNumberOfComments}
    \stepcounter{SSNumberOfComments}
    \newcommand{\santi}[1]{\textcolor{red}{\small  [SS\#\arabic{SSNumberOfComments}\stepcounter{SSNumberOfComments}: #1]}}
\else
    \newcommand\mc[1]{}
    \newcommand\aj[1]{}
    \newcommand\jv[1]{}
    \newcommand\santi[1]{}
\fi

\def\hb{\hbox to 11.5 cm{}}

\begin{document}

\pagestyle{headings}
\def\thepage{}

\begin{frontmatter}              

\title{Object Segmentation of Cluttered  Airborne LiDAR Point Clouds}

\markboth{}{Oct 2022\hb}

\author[A]{\fnms{Mariona} \snm{Carós}},
\author[B]{\fnms{Ariadna} \snm{Just}}
\author[A]{\fnms{Santi} \snm{Seguí}}
and
\author[A]{\fnms{Jordi} \snm{Vitrià}}

\runningauthor{M. Carós et al.}
\address[A]{Departament de Matemàtiques i Informàtica, Universitat de Barcelona (UB), \\ Gran Via Corts Catalanes, 585, 08007 Barcelona, Spain}
\address[B]{Institut Cartogràfic i Geològic de Catalunya, Barcelona, Spain}

\begin{abstract}
Airborne topographic LiDAR is an active remote sensing technology that emits near-infrared light to map objects on the Earth's surface. Derived products of LiDAR are suitable to service a wide range of applications because of their rich three-dimensional spatial information and their capacity to obtain multiple returns. However, processing point cloud data still requires a large effort in manual editing. Certain human-made objects are difficult to detect because of their variety of shapes, irregularly-distributed point clouds, and low number of class samples. In this work, we propose an efficient end-to-end deep learning framework to automatize the detection and segmentation of objects defined by an arbitrary number of LiDAR points surrounded by clutter. Our method is based on a light version of PointNet that achieves good performance on both object recognition and segmentation tasks. The results are tested against manually delineated power transmission towers and show promising accuracy.


\end{abstract}

\end{frontmatter}
\markboth{Oct 2022\hb}{Oct 2022\hb}



\section{Introduction}

\begin{figure*}[t]
\hfill
\includegraphics[width=\textwidth]{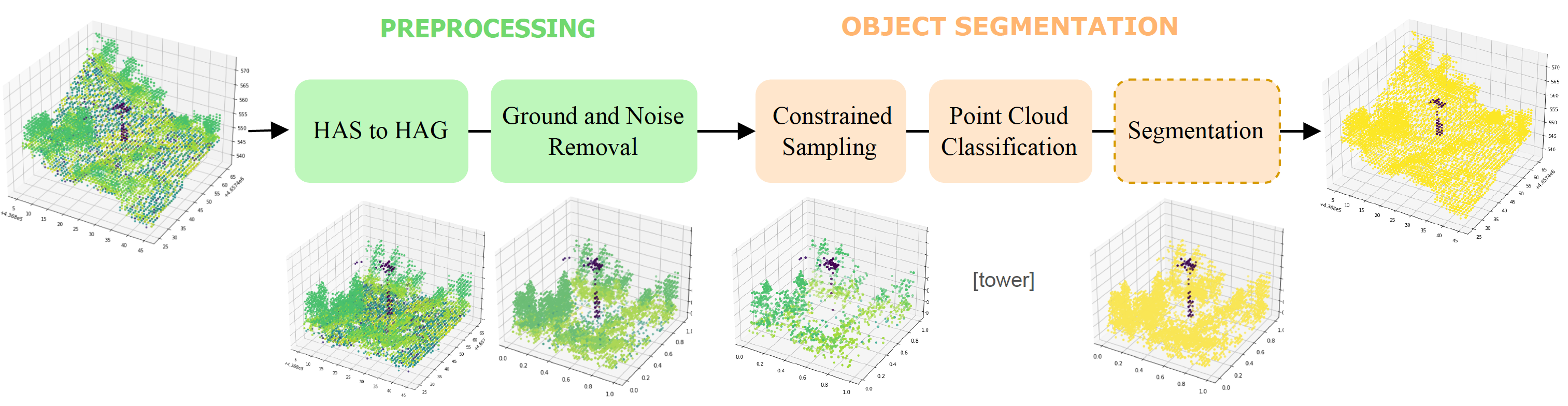}
\caption{Overview of our preprocessing and segmentation pipelines. Point heights are transformed from heights above sea (HAS) into heights above ground (HAG). Then, the point cloud is filtered from noise, constrained sampling is applied, and the resulting point cloud is used to train both the classifier and segmentation networks. During inference, the classification network detects if the object is within the processing point cloud, if it does, segmentation is applied to all input points. Otherwise, the block is skipped and all points are labeled as background.}
\label{fig:framework}
\end{figure*}

Light Detection and Ranging (LiDAR) is a technology that emits pulses of light to measure the distance from the sensor to the objects. Topographic LiDAR sensors generate pulses of near-infrared light that are reflected on the Earth's surface to create high-resolution three-dimensional (3D) maps of the surrounding environment. The data can be used for a variety of purposes, such as 
environmental monitoring \cite{r1}, forest inventories \cite{r6} or object detection \cite{r10}. 

Laser scanning can be mounted on different platforms depending on the target purpose. For indoor mapping or architecture heritage, Terrestrial Laser Scanning (TLS) is used, which is commonly placed on tripods. For road and 3D urban mapping, Mobile Laser Scanning (MLS) is the choice, usually mounted on vehicles. For large-scale applications including terrain modeling, forestry, and urban mapping, an Airborne Laser Scanning (ALS) is generally used; and for global monitoring of terrain and vegetation, LiDAR is installed in a satellite (SLS).


LiDAR observations are stored as point clouds, which are collections of points defined by 3D coordinates and may include features like intensity, number of return, or incidence angle.  Recent airborne LiDAR systems are hybrid in the sense they include an RGB and Near-Infrared (NIR) camera, capturing simultaneous images and assigning 4 channels of color to each LiDAR point cloud. The advancements of this 3D sensing technology are increasing the demand for point cloud processing techniques. Simultaneously, given the highly-accurate 3D information,  point cloud classification and segmentation have become an active research direction in the fields of remote sensing and computer science. There have been remarkable advances in deep learning techniques for point cloud understanding such as PointNet \cite{r12}, PointNet++\cite{r13}, RSNet \cite{r16}, Point Graphs \cite{r15}, GSPN \cite{r14} or Point Transformer \cite{r17}, and many of these approaches achieve impressive results. However, almost all of them are limited to synthetic data \cite{r8} or indoor datasets \cite{r7} and are difficult to be directly extended to real large-scale outdoor airborne LiDAR data, where the point clouds have significantly different number and distribution of points.
When dealing with automated 3D mapping with airborne LiDAR data, certain human-made objects are difficult to detect because of their irregularly-distributed point clouds and variety of shapes. Partial observations of objects are common due to occlusions, and background is blended with objects due to clutter in the real-world scenes. In addition, the low number of class samples makes it difficult to use deep learning models.

In order to overcome these challenges, we present an efficient end-to-end framework for object recognition and segmentation of LiDAR data and describe a real application case study. The proposed approach consists of a preprocessing stage with a sliding window to split our point cloud data into cubes, and a deep learning framework based on PointNet \cite{r12} to detect if the object is within the cube and segment it. A general overview of the framework is shown in Figure \ref{fig:framework}.

Specifically, in this work we make the following main contributions. First, we describe an end-to-end approach for processing large-scale airborne LiDAR data with deep learning. Second, we propose a 3D detection framework capable of segmenting a specific object defined by an arbitrary number of points. Third, we report several experiments on our  manually delineated airborne LiDAR dataset containing power transmission towers, where we conduct controlled studies to examine specific choices and parameters. Our code is publicly available at 
\href{https://github.com/marionacaros/3D-object-segmentation/tree/master}{https://github.com/marionacaros/3d-object-segmentation}. \mc{fer public el codi}

This article is structured as follows: Section \ref{sec:related-work} overviews literature and related work
in regards to 3D data modeling. Section \ref{sec:approach} presents the proposed method. The dataset is explained in section \ref{sec:dataset}. In section  \ref{sec:experiments} we report experiments and results. Finally, section \ref{sec:conclusions} summarizes the main conclusions.



\section{Related Work}
\label{sec:related-work}
Deep learning on 3D data has been receiving increasing attention in recent years. A number of different representations have been explored, including multi-view projections, voxel grids, and point clouds.

LiDAR data are irregularly distributed, unordered and scattered. Considering the success of Convolutional Neural Networks (CNNs) for image understanding, an intuitive approach is to project volumetric data into 2D planes \cite{r21}. Then, CNNs are used to extract feature representations from each of the planes. Nevertheless, these multi-view projections collapse spatial information of points which may affect object recognition. An alternative approach to use the benefits of CNNs on LiDAR data is to convert raw point cloud data into voxelized grids \cite{r11}. The irregular distribution of data is transformed into a uniform grid where 3D CNNs are applied. Compared to multi-view methods, this strategy has no loss of information, but can be very expensive in terms of computational and memory costs due to the generated number of voxels.

Rather than projecting irregular point clouds onto regular grids, point-based networks directly process point clouds. PointNet \cite{r12} was the pioneer work to directly process point sets. The key idea of PointNet is to process points independently by using permutation-invariant operators, and then aggregate them into a global feature representation by max-pooling. 
In the following work of these authors, PointNet++\cite{r13}, they incorporate local dependencies and hierarchical feature learning in the network to increase sensitivity to local geometric layout. A number of works relate point clouds to graphs and perform graph convolutions for feature extraction \cite{r15}. RSNet \cite{r16} uses Recurrent Neural Network (RNN) and a slice unpooling layer to project features of unordered points onto an ordered sequence of feature vectors. Point Transformer \cite{r17} exploits the positional information of points and applies self-attention to point clouds for semantic scene segmentation and object classification.

In this work, we select PointNet architecture as the base of our framework due to its simplicity, robustness and low execution time \cite{r19}. The goal of our work is to provide an end-to-end framework for real-world LiDAR applications. Thus, we not only focus on accuracy but memory consumption and training time as well.


\section{Proposed Approach}
\label{sec:approach}

Given a point cloud and a candidate label set, our task is to assign each of input points with one of the semantic labels in order to identify a specific object. Our method consists of two main stages presented in Figure \ref{fig:framework}: Preprocessing and Object Segmentation. Preprocessing blocks make a transformation to the data to simplify the segmentation task.

\subsection{Preprocessing}

The first step of our algorithm, in both train and inference tasks, is to partition the point cloud into smaller cubes, which enables parallelization and makes the segmentation task easier for the model. At the end of the pipeline prediction of cubes are merged again in the same point cloud. The size of the cubes is chosen considering two factors: The size of the object to be detected, and the minimum distance between objects. 

Once we have our reduced point clouds, we transform heights above sea (HAS) into heights above ground (HAG). Shapes may not be easily compared in hilly or mountainous terrain, because part of the observed variability is due to changes in the altitude of the surface. For this reason, we normalize all heights by subtracting ground's height\footnote{Topographic ground map is provided by an external source \cite{r32}} to $z$ coordinate values and obtain $z_{HAG}$. Next, we remove ground points ($z_{HAG}=0$) and noise (i.e. points above 100 meters). We consider that these types of points provide no valuable information to identify our target object, and by discarding them we increase the efficiency and performance of our model. Finally, we normalize them into a unit sphere.


\subsection{Object Segmentation}

The goal of this stage is to segment the target object within the point cloud window. The methodology is the following; The input point cloud is sampled to a fixed number of points, then a binary classifier is used to detect if the object is within the point cloud. If the result is negative, all points are labeled as background and the subsequent window is processed. On the contrary, the whole point cloud is fed into the segmentation model, which assigns a label to each point.

\begin{figure*}[t]
\hfill
\includegraphics[width=\textwidth]{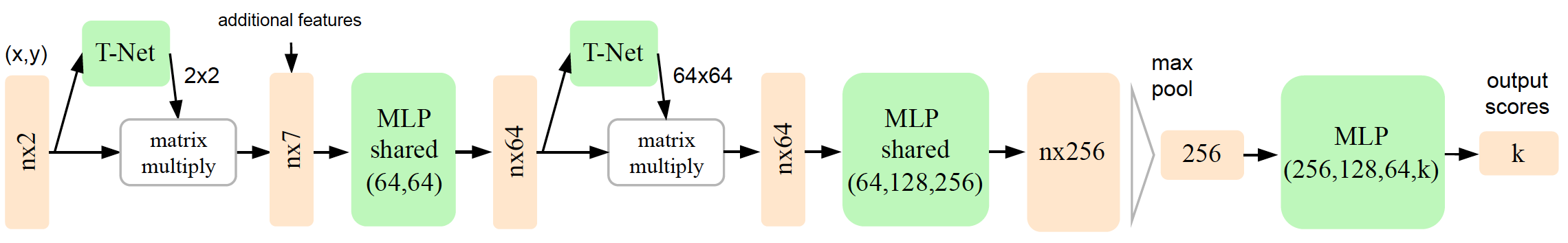}
\caption{The classification network follows the PointNet structure. It takes $(x,y)$ values of $n$ points as input, applies a transformation learned by T-Net, and it concatenates additional features to the output; such as $z$, intensity, and RGB.  Then, another transformation is applied to the feature space. Finally, features are aggregated by max pooling. The output is classification scores for k classes. MLP stands for multi-layer perceptron.}
\label{fig:net}
\end{figure*}

The algorithm behind both of our models is a lighter version of PointNet, where the number of parameters is reduced to one quarter. The classification network is shown in Figure \ref{fig:net}. We input $x$ and $y$ coordinates into the transformation net (T-Net) which learns a canonical representation of these 2 dimensions. Then, $z$ and additional features such as intensity, RGB, and NIR are concatenated. Notice that $z$ is not input into the canonical transformation because we do not expect our target objects to be tilted. Finally, features are aggregated by max pooling. The output of the network is the classification score, we set $k=2$ for object detection.

\begin{figure}[t]
\centering
\begin{subfigure}[b]{0.31\textwidth}
  \includegraphics[width=\linewidth]{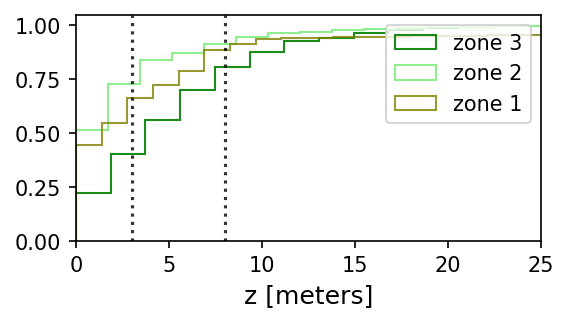}
  \caption{CDF of all $z_{HAG}$}
  \label{fig:cdf_all_p}
\end{subfigure}%
\begin{subfigure}[b]{.31\textwidth}
  \includegraphics[width=\linewidth]{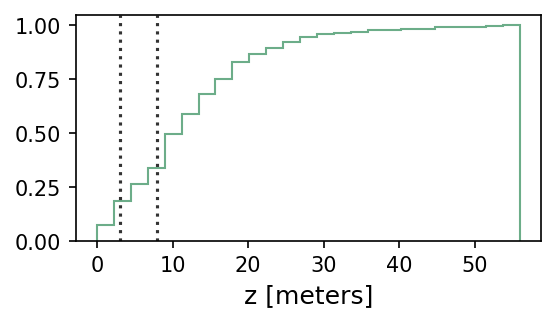}
  \caption{CDF of towers $z_{HAG}$}
  \label{fig:cdf_towers}
\end{subfigure}
\begin{subfigure}[b]{0.31\textwidth}
  \includegraphics[width=\linewidth]{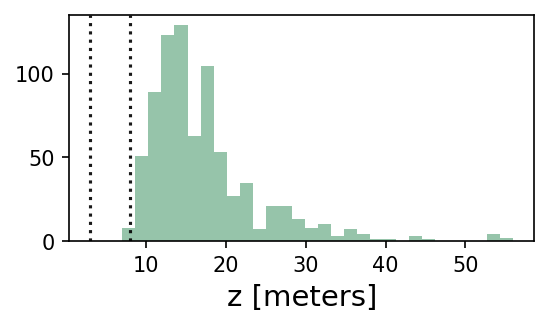}
  \caption{Towers heights}
  \label{fig:hist_heights}
\end{subfigure}
\caption{Cumulative distribution function (CDF) of points heights. Dotted lines identify the values of 3 and 8 meters, which correspond to low and medium vegetation in Catalonia.} 
\label{fig:hists_sampling}
\end{figure}

Training PointNet requires all input point clouds to be the same size. However, in LiDAR data the number of points can variate by tens of thousands. A usual approach is to randomly sample points from the point cloud and input them into the model in batches of a fixed size. A major problem of this procedure is the sampling of high-dense point sets, where we see a drop in performance caused by incomplete objects. This occurs when objects are small in comparison with clutter and random sampling causes points of the same object to be scattered among batches. To alleviate the aforementioned problem, we propose the {\bf Constrained Sampling} block, which takes advantage of the spatial location of points. We know that most of our dataset points correspond to vegetation, so we design a constrained sampling based on heights. In Figure \ref{fig:hists_sampling} we present three heights distributions obtained from point cloud windows containing our target object (a power transmission tower). 
From left to right we see the cumulative distribution function (CDF) of all $z_{HAG}$ values, $z_{HAG}$ of points labeled as a tower, and maximum $z_{HAG}$ per tower. By comparing Figures \ref{fig:cdf_all_p} and \ref{fig:cdf_towers} we notice a big difference between distributions of points. Regarding all $z_{HAG}$ values, there are between $40\%-75\%$ of the points in the range of $3-8$ meters, while the percentage is reduced to half if we focus on points labeled as tower. These values make total sense as 3 and 8 meters correspond to low and medium vegetation in Catalonia, while the minimum height of our target object is 10 meters (shortest tower in Figure \ref{fig:sub_towers}). 
In consideration of this, our sampling methodology is the following; We first down-sample points below 3 meters, if the number of points is larger than the defined fixed amount, we down-sample again raising the threshold to 8 meters. Finally, if we still have too many points, we randomly sample the whole point cloud. The goal of our sample strategy is to remove cluttered points and make our target object more visible to the model.

We apply constrained sampling as a previous step to both of our models in the training phase, so all input data is sampled according to this strategy. We compare the obtained results with random sampling in Section \ref{sec:experiments} and show that applying constrained sampling results in high precision results. During inference, all input points are fed into the segmentation model.


\section{Dataset}
\label{sec:dataset}

\begin{table}[b]
    \caption{Dataset properties}
    \label{tab:dataset}
    \begin{tabular}{c|ccc}
    \hline
                                    & \textbf{Zone 1} & \textbf{Zone 2}        & \textbf{Zone 3}       \\ \hline
    \textbf{Sensor properties}             & Intensity       & Intensity, RGB, NIR     & Intensity, RGB, NIR \\
    \textbf{Mean density $[pts/m^2]$} & 6               & 10                      & 8                  \\
    \textbf{Blocks size $[Km]$}       & $0.5 \times 0.5$       & $1 \times 1$                  &  $2 \times 2$                  \\ 
    \textbf{Total number of blocks}                   & 12              & 112                    &  27                  \\ \hline
    \textbf{Point cloud windows containing towers}  & 18           & 296                    & 475                   \\ 
    \textbf{Point cloud windows without towers}  & 2,044           & 53,756                 & 38,296 \\ \hline
    \end{tabular}
\end{table}

Our dataset is composed of three large-scale outdoor areas from several ALS flights in Catalunya, each covering 3, 112 and 108 squared kilometers (total of 223 squared kilometers). These data were collected with two different LiDAR sensors by ICGC\footnote{Institut Cartogràfic i Geològic de Catalunya}. The first one is a Leica ALS50-II, which only allows capturing signal intensity data. The second one is a Terrain Mapper 2, which combines a LiDAR sensor with two nadir cameras in RGB and NIR. Each point is defined by $xyz$ coordinates, intensity, RGB, and NIR. The obtained mean point density varies with each flight and sensor, from 6 to 10 $pts/m^2$, as it is shown in Table \ref{tab:dataset}. 

Points are labeled into four classes: Power transmission tower, power lines,  ground, and background. The latter includes all the points different from the previous classes. In order to train a model for power tower recognition, we need each tower in a separate point cloud. These point cloud cubes are obtained by using a sliding window approach that we detail next.

\begin{figure}
\centering
\begin{subfigure}[b]{0.19\textwidth}
  \includegraphics[width=\linewidth]{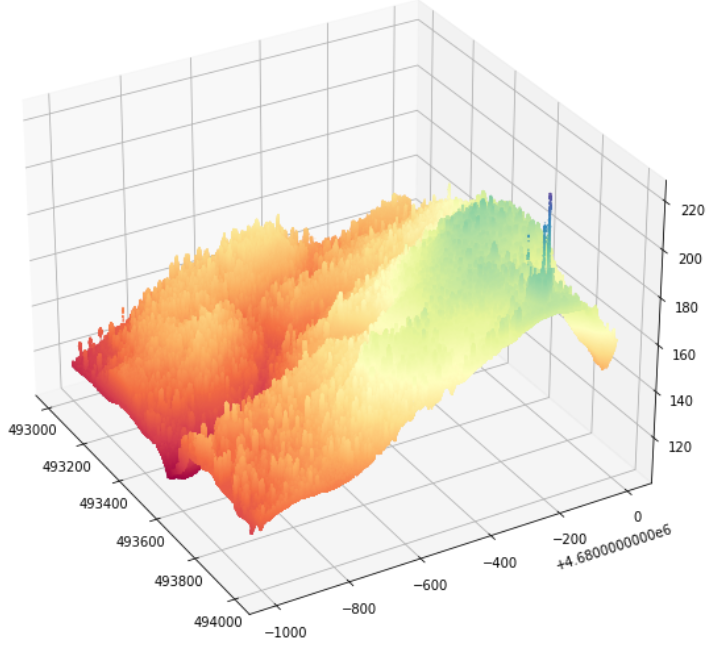}
  \caption{Point cloud}
  \label{fig:sub1}
\end{subfigure}%
\begin{subfigure}[b]{.19\textwidth}
  \includegraphics[width=\linewidth]{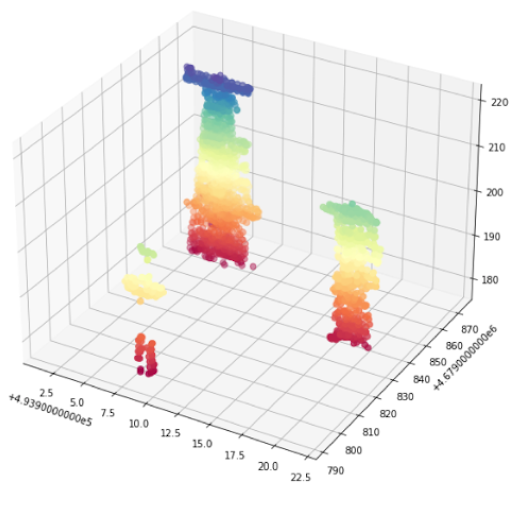}
  \caption{Towers}
  \label{fig:sub2}
\end{subfigure}
\begin{subfigure}[b]{0.19\textwidth}
  \includegraphics[width=\linewidth]{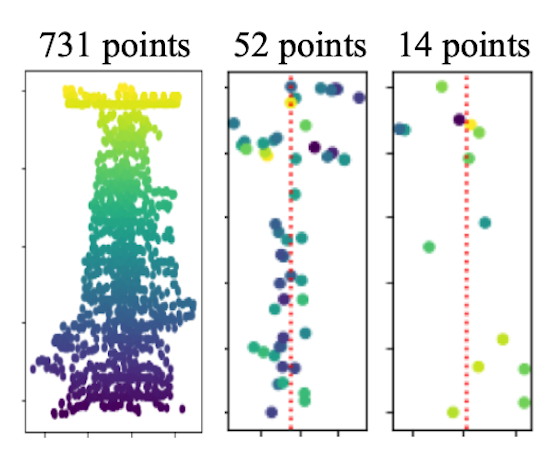}
  \caption{Splitting}
  \label{fig:sub3}
\end{subfigure}
\begin{subfigure}[b]{.19\textwidth}
  \includegraphics[width=\linewidth]{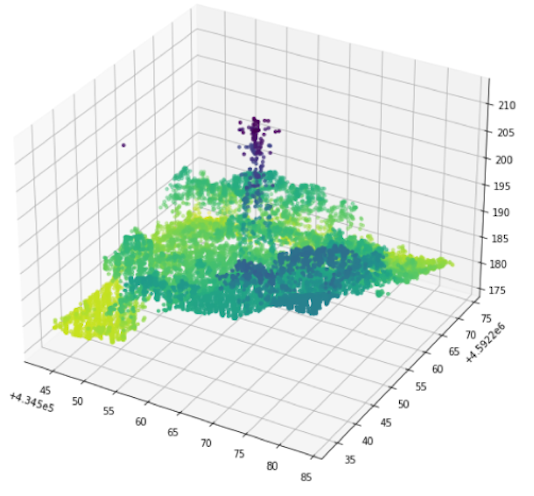}
  \caption{Tower + BG}
  \label{fig:sub4}
\end{subfigure}
\begin{subfigure}[b]{.19\textwidth}
  \includegraphics[width=\linewidth]{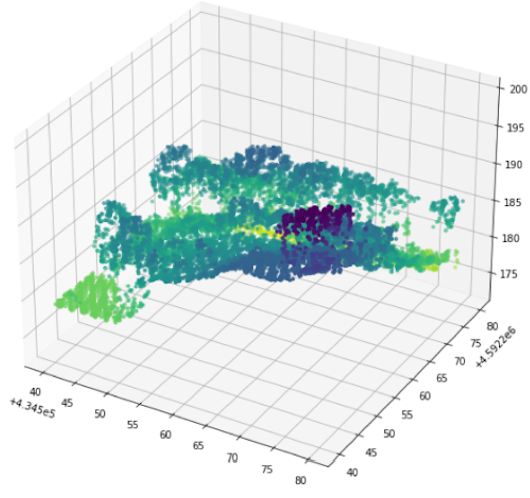}
  \caption{Only BG}
  \label{fig:sub5}
\end{subfigure}
\caption{Dataset processing pipeline for training data. From left to right: (a) Input point cloud cube of 1 $\times$ 1 Km. (b) Points labeled as tower. (c) Separate tower point sets. (d) Tower with background points. (e) Only background points.}
\label{fig:test}
\end{figure}

\begin{figure}[b]
\centering
\begin{minipage}{.48\textwidth}
  \centering
  \includegraphics[width=0.9\linewidth]{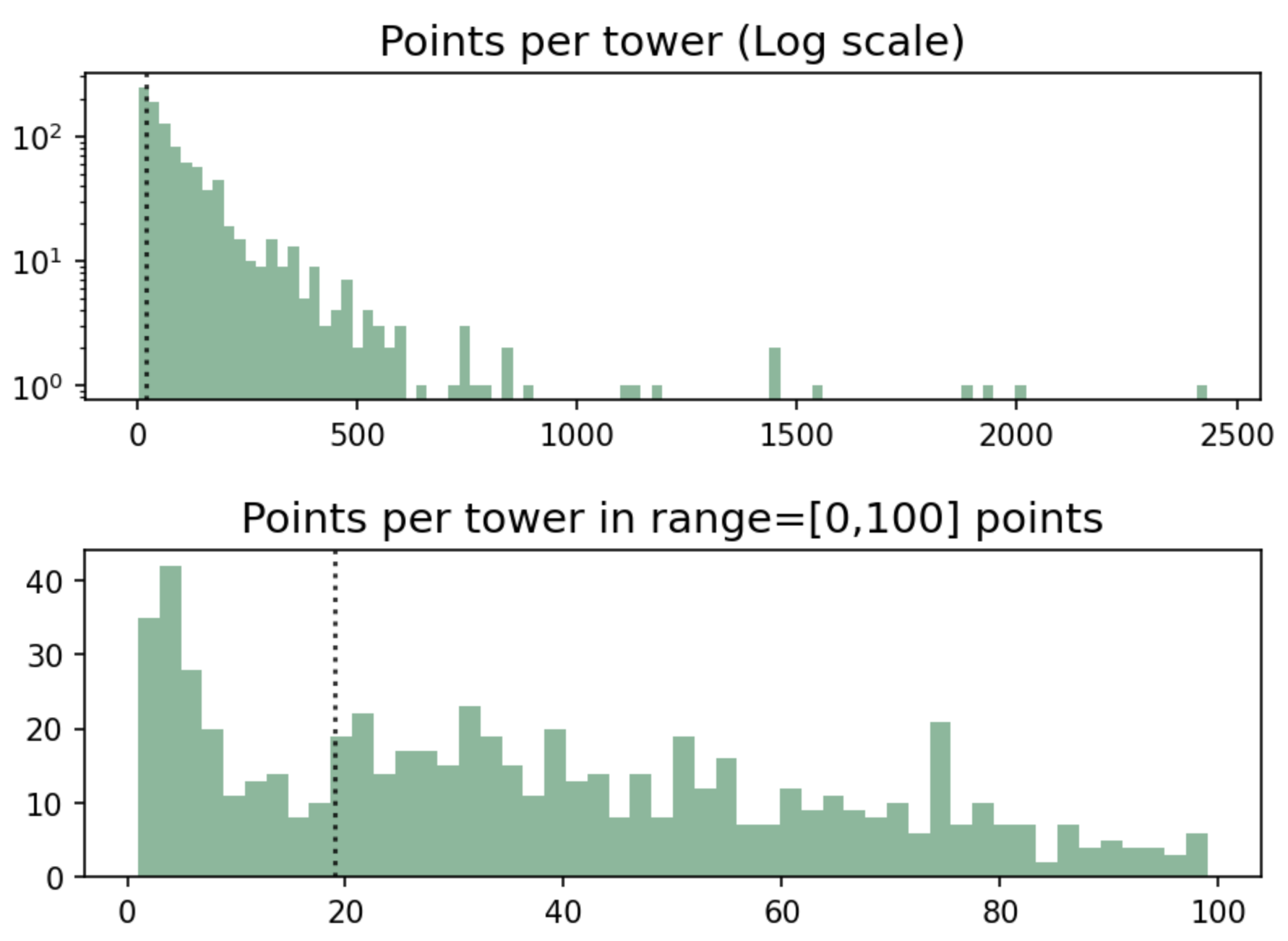}
  \captionof{figure}{Distribution of points per tower}
  \label{fig:sub_hist}
\end{minipage}%
\begin{minipage}{.5\textwidth}
  \centering
  \includegraphics[width=\linewidth]{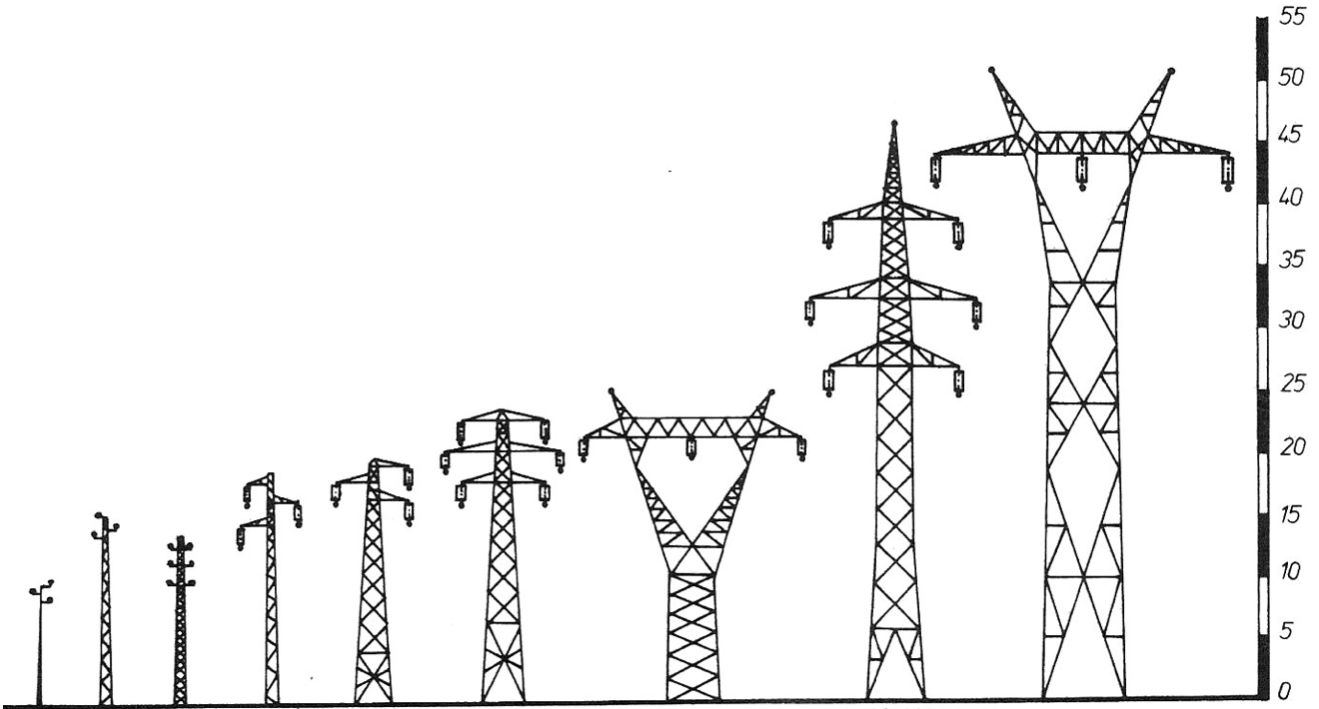}
  \captionof{figure}{Each type of tower has a specific shape and height. The highest tower is 50 meters and the shortest is 10 meters.}
  \label{fig:sub_towers}
\end{minipage}
\end{figure}

Given a LiDAR point cloud (Figure \ref{fig:sub1}), we identify points labeled as a power transmission tower and apply a fixed-size sliding window to split them on the $xy$ plane resulting in separate tower point sets. We use a window of $20 \times 20$ meters to avoid having two towers in the same point set, as the maximum tower's width in our dataset is 20 meters. Even though in some cases we find narrow towers sharing a window, these cases are very rare. Once towers are segmented (Figure \ref{fig:sub3}), we store those with more than 19 points, which represents 80\% of the towers. The number of points per tower is very variable, as presented in Figure \ref{fig:sub_hist}, ranging from 1 to 2434, and towers comprising less than 20 points do not present a clear shape. Some examples are illustrated in Figure \ref{fig:sub3}. Next, we get the center of each tower by computing the mean of $(x, y)$ coordinates. This location is used to get all points within the window, including cables and background. 
Finally, we normalize points in the unit sphere and store two versions of the same cube; the first one containing all types of points, and a second one with only ground and background points. This is done to facilitate the task of distinguishing between point clouds that have the target object and those that do not.

We end up with a total of 789 point cloud windows containing power transmission towers (see Table \ref{tab:dataset}), which represents only 0.84\% of the data. The dataset is highly imbalanced and considering that power transmission towers show a variety of shapes, as presented in Figure \ref{fig:sub_towers}, it is a challenging task to develop an accurate model for object recognition. Models trained on such data perform poorly for weakly represented classes \cite{r30}. However, by using appropriate preprocessing techniques and data augmentation, we achieve good results.

\section{Experimentation and Results}
\label{sec:experiments}

In this section, we report the performance of our classification and segmentation models in terms of accuracy and computational cost. We also study the effects of various processing and architecture choices with an ablation study.

\subsection{Implementation Details}


To address the problem of under-represented positive samples, we use two class balancing strategies: Data augmentation and a class-balancing weighted loss.

We augment the training set by storing a different $xy$ plane of each positive sample, which doubles the number of our target object samples. Specifically,  we add a random variation of 10 meters to each $x$ and $y$ to place towers at any part of the $xy$ plane within the point cloud. We use windows of 40 x 40 meters so large towers are not cut in half.

Regarding class-balancing, we weight the negative log-likelihood loss by using the effective number of samples presented in \cite{r28}. The effective number of samples is defined as the volume of samples and can be calculated by the following formula $(1 - \beta^n)/(1 - \beta)$, where $n$ is the number of samples and $\beta \in [0, 1)$ is a hyperparameter. 
The hyperparameter $\beta$ smoothly adjusts the class-balanced term between no re-weighting and re-weighing by inverse class frequency. We experiment with different $\beta$ values (0.9, 0.999, 0.9999), as the authors suggest, and compare their results.

We implement our code in Pytorch v1.8. with CUDA 11.6. We use Adam optimizer and an initial learning rate of 0.001 with 0.5 decay when loss does not decrease for 10 epochs. Models are trained for 50 epochs with an early stopping on the validation loss. The system used for the experiments has the following configuration: (i) CPU: Intel Xeon Silver 4210, (ii) RAM: 126GB, (iii) GPU: Quadro RTX 5000 - 16 GB, and (iv) OS: Ubuntu 20.04.

\subsection{Experimental Set-Up}

Each point is represented by a 7-dim vector of $xyz$ coordinates, intensity, green, blue, and Normalized Difference Vegetation Index (NDVI). NDVI is used in remote sensing \cite{r31} to indicate whether or not the target being observed contains live green vegetation, and it is computed by using a simple formula: $(NIR - red)/(NIR + red)$.

\iftrue
\begin{table}[t]
    \caption{Dataset train-test split}
    \label{tab:dataset_split}
    \begin{tabular}{c|ccc}
    \hline
                   & Zone 1 & Zone 2        & Zone 3       \\ \hline
    \textbf{Blocks with towers}   & 7              & 70                      &  21                  \\    
    \textbf{Blocks for train}       & 11              & 105                    &  24                  \\ 
    \textbf{Blocks for test}      & 1               & 7                       &  3                  \\  \hline
    \end{tabular}
\end{table}
\fi

The train-test split is done considering the number of blocks with towers. We set aside 10\% of blocks with towers for evaluation, which is reported in Table \ref{tab:dataset_split}, so that during training none of the point clouds belonging to the test blocks are observed. In classification, we use a training set of 80757 point clouds, a validation set of 6970, and a test set of 7935. During segmentation, background samples are reduced to 5\%, as they do not add value for learning the segmentation task. In training time we use 2048 points and a  batch size of 32, while in inference we use all points in batches of 1. For evaluation metrics, we use F$_1$ score in classification and mean classwise intersection over union (mIoU) in segmentation.

\subsection{Quantitative and Qualitative Results}

The results of classification and segmentation are presented in Tables \ref{tab:classification_res} and \ref{tab:seg_res}, respectively. In both models, the best results are achieved when using all available LiDAR features (intensity, RGB and NIR), together with constrained sampling and weighted loss with $\beta$ set to 0.999, which yields weights of [0.42, 0.58] in Classification and [0.4, 0.6] in Segmentation. The difference between weights is explained by the fact that in classification positives are defined by point clouds containing a tower, while in segmentation the positives are the points labeled as a tower.

\begin{figure*}[b]
\hfill
\includegraphics[width=\textwidth]{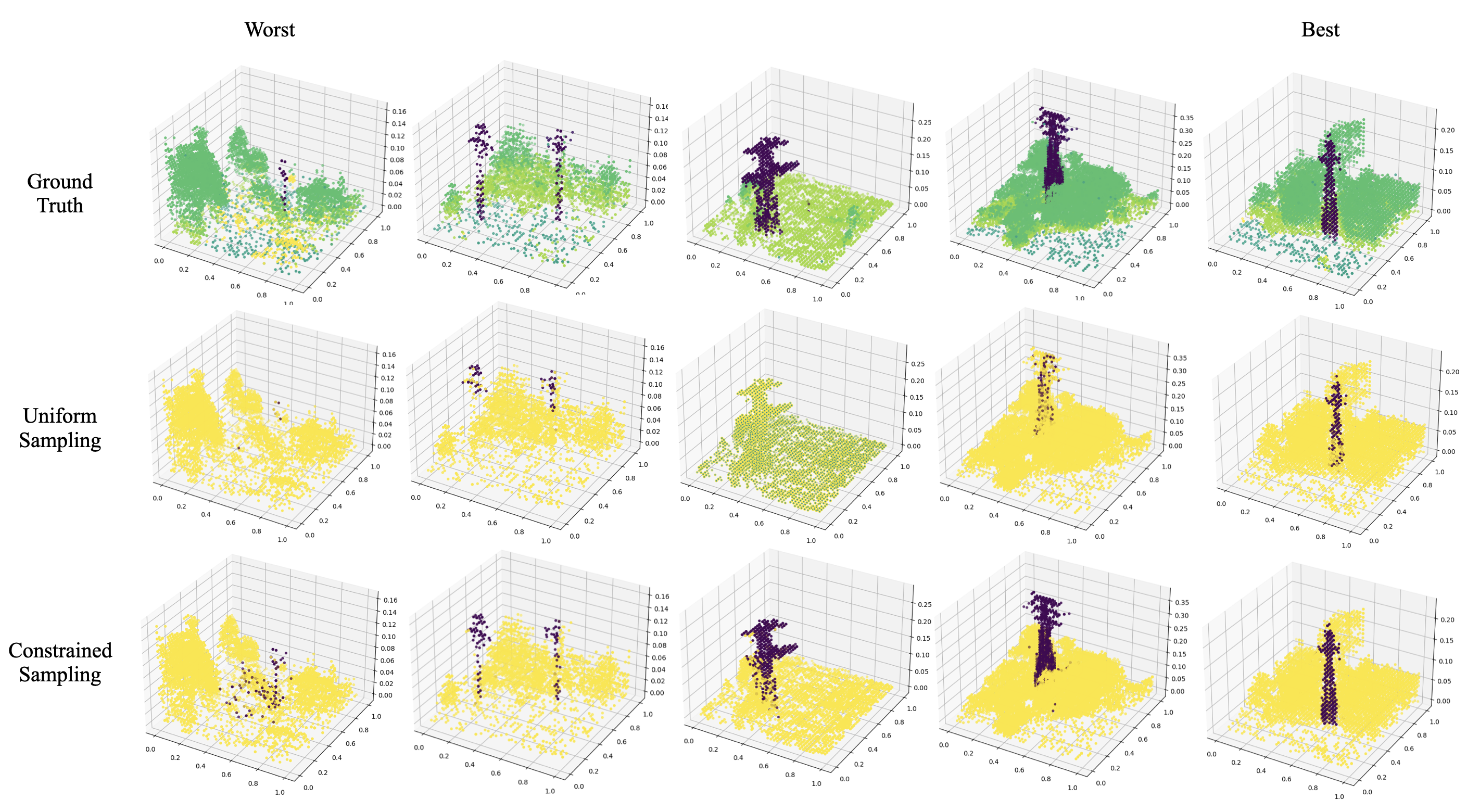}
\caption{Visualization for object segmentation results. Top row is the ground truth. Bottom and middle rows are results produced by our model with and without constrained sampling.}
\label{fig:segmen_results}
\end{figure*}

We qualitatively evaluate the segmentation capability of our method to unseen point clouds. Figure \ref{fig:segmen_results} shows the best and worst predictions using both uniform and constrained sampling. We can see that the predictions using constrained sampling are very close to the ground truth. Our model captures the whole tower shape when there is vegetation around, and in some cases, it is even able to detect two towers sharing the same point cloud window. When the object is not entirely in the cube some points are missed. Nevertheless, the majority of them are well predicted. The worst predictions are caused by small and low-density towers surrounded by high vegetation, as expected. 

\subsection{Ablation Study}

We now study the variation in performance involved in each specific decision. In particular, several key features and parameters are considered: (i) The use of RGB and NIR features; (ii) the use of constrained sampling; (iii) the effect of class-balancing weights; and (iv) the amount of trainable parameters. Results are shown in Tables \ref{tab:classification_res} and \ref{tab:seg_res}.

\textbf{RGB and NIR.} We first investigate the effect of using color and near-infrared information. We can see that without using these extra features the performance of both models drops significantly, being 8.3 absolute percentage points in classification  and 22.8 in segmentation. This suggests that RGB and NIR are essential for a proper object segmentation.

\textbf{Constrained sampling.} We study the type of sampling applied to the point clouds.  When using random sampling both models perform worst, the models may not be able to learn the shape of the object because of all the clutter in the scene. The implementation of constrained sampling results in an improvement of 2 absolute percentage points in classification ($F_1$ score), as well as an increase of 4.9 in the mean Intersection over Union (mIoU) in segmentation. Additionally, the IoU score for tower segmentation shows a substantial improvement of 10.5 which is visually noticed in Figure~\ref{fig:segmen_results}.

\textbf{Class-balancing weights.} We conduct an ablation study of $\beta$. When $\beta$ is set to 0.9, both classes are assigned the same weight and the model miss to detect a lot of positives. When $\beta$ is set to 0.9999, the predictions result in a lot of false positives. The best performance is achieved when $\beta$ is set to 0.999. 

\textbf{Trainable parameters.} We compare the performance between the original PointNet and our lighter version (PointNet$_L$), where the number of parameters is reduced from 3.5M to 0.9M. We see that PointNet$_L$ is not only more efficient but outperforms PointNet in terms of precision. This is probably caused by having too many parameters to learn.


\begin{table}[t]
\caption{Results of classification and ablation study (features, sampling, and weights) sorted by $F_1$ score. Subscript \textit{L} denotes \textit{light} version of PointNet. Sampling* refers to constrained sampling. Time and memory consumption during inference of the test set are also reported.}
\label{tab:classification_res}
\begin{tabular}{c|cccc|cc|c}
\hline
Model           & RGB+NIR & Sampling* & $\beta$ & Weights      & Time $[h]$ & Mem. $[GiB]$ & $F_1$ score \\\hline
PointNet$_L$   & x      & x       & 0.9999   & [0.1, 0.9]     & 3.0      & 2.25 & 78.4          \\\hline
PointNet$_L$   &        & x       & 0.999   & [0.42, 0.58]    & 1.6     & 2.25 & 84.4          \\\hline
PointNet        & x      & x     & 0.999    & [0.42, 0.58]    & 3.4    & 6.7 & 88.3         \\ \hline
PointNet$_L$   & x      & x       & 0.9   & [0.5, 0.5]      & 1.5     & 2.25 & 89.7          \\\hline
PointNet$_L$   & x      &         & 0.999   & [0.42, 0.58]    & 2.1     & 2.25 & 90.7          \\\hline
PointNet$_L$   & x      & x       & 0.999   & [0.42, 0.58]    & 3.0     & 2.25 & \textbf{92.7} \\\hline
\end{tabular}
\end{table}

\begin{table}[b]
\caption{Results of segmentation and ablation study (features, sampling, and weights) sorted by mIoU. Subscript \textit{L} denotes \textit{light} version of PointNet. Sampling* refers to constrained sampling.}
\label{tab:seg_res}
\begin{tabular}{c|ccc|cc|c}
\hline
Model & RGB+NIR & Sampling* & Weights  & IoU tower & IoU veg & mIoU  \\ \hline
PointNet$_L$   &        & x       & [0.4, 0.6]      & 28.0     & 98.0   & 63.0          \\ \hline
PointNet$_L$   & x      &         & [0.5, 0.5]       & 32.9     & 98.4   & 65.7          \\ \hline
PointNet$_L$   & x      & x        & [0.5, 0.5]       & 55.0     & 98.7   & 76.9          \\ \hline
PointNet$_L$   & x      &         & [0.4, 0.6]      & 63.0     & 98.8   & 80.9          \\ \hline
PointNet       & x      & x       & [0.4, 0.6]       & 71.7     & 96.6   & 84.2          \\ \hline
PointNet$_L$   & x      & x       & [0.4, 0.6]      & 73.5     & 98.0   & \textbf{85.8} \\ \hline
\end{tabular}
\end{table}

\section{Conclusions}
\label{sec:conclusions}

We proposed an efficient object segmentation method for point cloud data that is going to be operationally implemented in ICGC's production lines. 
Our approach enables the segmentation of objects defined by variable point sets given a large outdoor scene. We report several experiments on our manually delineated airborne LiDAR dataset, where we conduct controlled studies to examine specific choices and parameters. We conclude that color and near-infrared features are essential for proper object segmentation, constrained sampling is key in cluttered point clouds, and the selection of class-balancing weights is important when dealing with imbalanced datasets to achieve a good performance.

\section{Acknowledgments}

This research was funded by an industrial doctorate grant of AGAUR between Universitat de Barcelona and Institut Cartogràfic I Geològic de Catalunya. This work was partially funded by projects RTI2018-095232-B-C21 (MINECO/FEDER, UE) and 2017SGR1742 (Generalitat de Catalunya).


\begin{thebibliography}{99}

\bibitem{r1}
Almeida, Danilo Roberti Alves de, et al. "The effectiveness of lidar remote sensing for monitoring forest cover attributes and landscape restoration." Forest Ecology and Management 438 (2019): 34-43.

\bibitem{r6}
Michałowska, Maja, and Jacek Rapiński. "A review of tree species classification based on airborne LiDAR data and applied classifiers." Remote Sensing 13.3 (2021): 353.

\bibitem{r7}
Armeni, Iro, et al. "3d semantic parsing of large-scale indoor spaces." Proceedings of the IEEE CVPR conference, 2016.

\bibitem{r8}
Chang, Angel X., et al. "Shapenet: An information-rich 3d model repository." 2015.

\bibitem{r10}
Zhou, Yin, et al. End-to-end multi-view fusion for 3d object detection in lidar point clouds. Conference on Robot Learning. PMLR, 2020.

\bibitem{r11}
Wu, Zhirong, et al. 3d shapenets: A deep representation for volumetric shapes. Proceedings of the IEEE CVPR conference, 2015. 

\bibitem{r12}
Qi, Charles R., et al. "Pointnet: Deep learning on point sets for 3d classification and segmentation." Proceedings of the IEEE  CVPR conference, 2017.

\bibitem{r13}
Qi, Charles Ruizhongtai, et al. "Pointnet++: Deep hierarchical feature learning on point sets in a metric space." Advances in neural information processing systems 30, 2017.

\bibitem{r14}
Yi, Li, et al. "Gspn: Generative shape proposal network for 3d instance segmentation in point cloud." Proceedings of the IEEE CVPR conference, 2019.

\bibitem{r15}
Landrieu, Loic, and Martin Simonovsky. "Large-scale point cloud semantic segmentation with superpoint graphs." Proceedings of the IEEE CVPR conference, 2018.

\bibitem{r16}
Huang, Qiangui, Weiyue Wang, and Ulrich Neumann. "Recurrent slice networks for 3d segmentation of point clouds." Proceedings of the IEEE CVPR conference, 2018.

\bibitem{r17}
Zhao, Hengshuang, et al. "Point transformer." Proceedings of the IEEE/CVF ICCV conference, 2021.

\bibitem{r19}
Zoumpekas, T., et al. Benchmarking Deep Learning Models on Point Cloud Segmentation. Proceedings of the 23rd International Conference of the Catalan Association for Artificial Intelligence, 2021.

\bibitem{r21}
Lang, Alex H., et al."Pointpillars: Fast encoders for object detection from point clouds." Proceedings of the IEEE/CVF CVPR Conference, 2019.

\bibitem{r28}
Cui, Yin, et al. "Class-balanced loss based on effective number of samples." Proceedings of the IEEE/CVF CVPR conference, 2019.

\bibitem{r30}
Van Horn, Grant, et al. "The devil is in the tails: Fine-grained classification in the wild.", 2017.

\bibitem{r31}
Pettorelli, N. The normalized difference vegetation index. Oxford University Press, 2013.

\bibitem{r32}
Terrasolid software. https://geocue.com/software/terrasolid/


\end{thebibliography}
\end{document}